\documentclass{ctr_summer}

\usepackage{ctrfont}
\usepackage{natbib}
\usepackage{undertilde}
\usepackage{color}

\usepackage{graphicx}

\usepackage{amsmath,rotating}
\usepackage{dashrule}
\usepackage{undertilde}

\usepackage{url}

\usepackage{color}

\let\square\undefined

\usepackage{amsmath}
\usepackage{amssymb}

\DeclareMathOperator*{\argmin}{arg\,min}

\newcommand\restr[2]{{
  \left.\kern-\nulldelimiterspace 
  #1
  \vphantom{|}
  \right|_{#2}
}}

\title{Proper latent decomposition}

\shorttitle{Proper latent decomposition}

\author{D. Kelshaw\footnote[1]{Department of Aeronautics, Imperial College London, United Kingdom\label{fn:refnote}} \and L. Magri\footref{fn:refnote}\footnote[2]{The Alan Turing Institute, United Kingdom}}

\shortauthor{Kelshaw \& Magri}

\begin{document}

%% Setting the first page number.  Please leave as it is, at "1".
\setcounter{page}{1}
\maketitle

%% ABSTRACT: 
In this paper, we introduce the proper latent decomposition (PLD) as a generalization of the proper orthogonal decomposition (POD) on manifolds. PLD is a nonlinear reduced-order modeling technique for compressing high-dimensional data into nonlinear coordinates.
First, we compute a reduced set of intrinsic coordinates (latent space) to accurately describe a flow with fewer degrees of freedom than the numerical discretization. The latent space, which is geometrically a manifold, is inferred by an autoencoder.
Second, we leverage tools from differential geometry to develop numerical methods for operating directly on the latent space; namely, a metric-constrained Eikonal solver for distance computations. With this proposed numerical framework, we propose an algorithm to perform PLD on the manifold.
Third, we demonstrate results for a laminar flow case and the turbulent Kolmogorov flow. 
For the laminar flow case, we are able to identify a semi-analytical expression for the solution of Navier-Stokes; in the Kolmogorov flow case, we are able to identify a dominant mode that exhibits physical structures, which are compared with POD.
This work opens opportunities for analyzing autoencoders and latent spaces, nonlinear reduced-order modeling and scientific insights into the structure of high-dimensional data. \\

\hrule

%% SECTION NAMES:  
%% Should be lower-case except for proper nouns and abbreviations
%% Should not end with a period
%%%%%%%%%%%%%%%%%%%%%%%%%%%%%%%%%%%%%%%%%%%%%%%%%%%%%%%%%%%%%%%%%%%%%%%%%%%%%%%%
\section{Introduction} \label{sec:introduction}
%%%%%%%%%%%%%%%%%%%%%%%%%%%%%%%%%%%%%%%%%%%%%%%%%%%%%%%%%%%%%%%%%%%%%%%%%%%%%%%%

Turbulent flows are notoriously challenging to model, which is a consequence of nonlinearities in the Navier-Stokes equations. Typical predictive strategies rely on discretizations with large number of degrees of freedom, leading to a high-dimensional state space; this space may be prohibitively large for accurate analysis of system dynamics and statistics. However, because turbulent flows are dissipative, once the transient dynamics have decayed, the turbulent dynamics converge to an attractor. This attractor typically occupies a relatively small portion of the entire phase space and constitutes a nonlinear manifold. The objective of this paper is to identify an approximate set of intrinsic coordinates, describing the turbulent attractor in a lower-dimensional state space, which we refer to as nonlinear reduced-order modeling.

Many of the current methods for reduced-order modeling are  linear, such as proper orthogonal decomposition (POD)~\citep{chatterjee2000introduction} and dynamic mode decomposition \citep{schmid2009dynamic}. These methods provide good compression for periodic or quasi-periodic systems, but performance may degrade as a consequence of the chaotic dynamics observed within turbulent flows. The inherent, linear nature of these methods amount to model assumptions and simplifications, which may result in systematic model error. Employing a nonlinear reduced-order modeling method seeks to alleviate these issues. Machine learning methods for nonlinear reduced-order modeling are prevalent, with autoencoders providing a means by which to obtain a nonlinear compressed representation of the state space of turbulence~\citep{eivazi2020deep, maulik2021reduced}. Analytically, it is possible to show that an autoencoder with purely linear activations is  equivalent to  POD on the data; however, in the presence of activations inducing nonlinearities the latent representation corresponds to intrinsic coordinates on a nonlinear manifold~\citep{magri2022InterpretabilityProperLatent}. The latent space discovered by an autoencoder, while compact, does not provide an interpretable representation of the system dynamics. Efforts to interpret autoencoders have been made, but typically employ ad-hoc methods~\citep{fukami2020convolutional, murata2020nonlinear}. Ideally, reduced-order models of turbulence should provide a parsimonious representation; that is, modes or intrinsic coordinates should directly correlate with identifiable physical phenomena present within the turbulent flow.

In this work, we propose the proper latent decomposition (PLD) as a nonlinear generalization of POD. We outline an approach in which a numerical representation of a latent manifold is extracted through an autoencoding approach, and geometric properties of the manifold are used to obtain principal geodesics that best describe the data. We draw on theoretical work from a previous Summer Program \citep{magri2022InterpretabilityProperLatent} and methods developed to operate on manifolds from differential geometry, as in \citet{kelshaw2024ComputingDistancesMeansb}, to achieve this.

The paper is structured as follows. In Section \ref{sec:review}, we provide a brief review of topics in differential geometry that we will use in our proposed methodology. Section \ref{sec:methodology:proper_latent_decomposition} introduces the notion of PLD at a high level, outlining key challenges in obtaining a latent representation, computing distances and operating directly on the manifold. We address the challenges of obtaining a nonlinear manifold in Section \ref{sec:methodology:autoencoders}, outline a differentiable method for computing distance in Section \ref{sec:methodology:distances}, and provide a framework for operating on the manifold in Section \ref{sec:methodology:log_map}. Finally, results are shown in Section~\ref{sec:results}, where we demonstrate PLD on a laminar wake trailing a bluff body and the two-dimensional turbulent Kolmogorov flow. We further discuss ongoing challenges in Section~\ref{sec:discussion}, before concluding in Section~\ref{sec:conclusions}.

%%%%%%%%%%%%%%%%%%%%%%%%%%%%%%%%%%%%%%%%%%%%%%%%%%%%%%%%%%%%%%%%%%%%%%%%%%%%%%%%
\section{Brief review of differential geometry} \label{sec:review}
%%%%%%%%%%%%%%%%%%%%%%%%%%%%%%%%%%%%%%%%%%%%%%%%%%%%%%%%%%%%%%%%%%%%%%%%%%%%%%%%

Differential geometry provides tools for the study of smooth manifolds, of which Euclidean spaces are a subset. In this work, we seek to obtain a nonlinear reduced-order manifold for describing turbulent flows; in order to work on such a manifold, we need to adopt differential geometry. In this section, we provide a brief overview of key concepts from differential geometry, which provide the basis of our proposed methodology. For a comprehensive overview of the subject, we refer the reader to~\cite{lee2018IntroductionRiemannianManifolds}.

\subsection{Riemannian manifolds, metrics and inner products} \label{sec:review:manifolds}
A Riemannian manifold is a pair $(M, g)$ in which $M$ is a smooth manifold \citep{lee2018IntroductionRiemannianManifolds} and $g: T_p M \times T_p M \rightarrow \mathbb{R}$ is a choice of Riemannian metric on $M$, where $T_p M$ denotes a vector in the tangent space at $p \in M$. Given vectors $v, w \in T_p M$, the metric defines the inner product on the tangent space~\citep{lee2018IntroductionRiemannianManifolds} as
\begin{equation} \label{eqn:review:metric}
	\langle v, w \rangle_g = g_p (v, w) = g_{ij} v^{i} w^{j},
\end{equation}
where components of the metric $g_{ij}(p) = \langle \restr{\partial_i}{p}, \restr{\partial_j}{p} \rangle$, and where in turn $\partial_i = \partial/\partial x^i$ are the basis vectors. Suppose $(\tilde{M}, \tilde{g})$ is a Riemannian manifold, and $M \subseteq \tilde{M}$ is an embedded submanifold. Given a smooth immersion $\iota: M \hookrightarrow \tilde{M}$, where $\dim M \leq \dim \tilde{M}$, the metric $g = \iota^\ast \tilde{g}$ is referred to as the metric induced by $\iota$, where $\iota^\ast$ is the pullback~\citep{lee2018IntroductionRiemannianManifolds}
\begin{equation} \label{eqn:review:induced_metric}
	g_p (v, w) = (\iota^\ast \tilde{g}) (v, w) = \tilde{g}_{\iota(p)} (d \iota_p (v), d \iota_p (w)).
\end{equation}

\subsection{Geodesics} \label{sec:review:geodesics}

A geodesic $\gamma: [a, b] \subset \mathbb{R} \rightarrow M$ is a locally length-minimizing curve, which generalizes the notion of a straight line on the manifold. These curves satisfy
\begin{equation} \label{eqn:review:geodesic}
	\nabla_{\dot{\gamma}} \dot{\gamma} = \ddot{\gamma}^k + \Gamma^k_{\phantom{k}ij} \dot{\gamma}^i \dot{\gamma}^j = 0,
		\quad \text{s.t.} \quad 
	\restr{\gamma}{\lambda = 0} = \gamma_0, \; 
	\restr{\dot{\gamma}}{\lambda = 0} = \dot{\gamma}_0, \; 
	\lambda \in [0, 1],
\end{equation}
where $\nabla_{v} w$ denotes the covariant derivative of a vector $w \in T_p M$ along a vector field $v \in TM$; Christoffel symbols $\Gamma^k_{\phantom{k}ij}$ are elements of the affine connection; and derivatives $\ddot{\gamma}, \dot{\gamma}$ are taken with respect to an affine parameter $\lambda \in \mathbb{R}$. A useful property of the geodesic equation is that it provides the means to map points on the manifold to a given tangent space, and vice versa. Given a point $p \in M$, we define $\exp_p: T_p M \rightarrow M, \log_p: M \rightarrow T_p M$, where $\exp_p \circ \log_p : q \mapsto q \; \forall q \in M$. This $\log$ map is not necessarily unique.

The notion of distance on the manifold is defined through geodesics. While all geodesics are length-minimizing, this is only in a local sense. Given points $p, q \in M$, the geodesic distance is defined as the infimum of the length of all valid geodesics~\citep{lee2018IntroductionRiemannianManifolds}
\begin{equation} \label{eqn:review:distance}
  d_g\left( p, q \right) = \inf_\gamma \left\{ \int_0^1 \langle \dot{\gamma}(t), \dot{\gamma}(t) \rangle^{0.5}_{\gamma(t)} dt : \gamma(0) = p, \gamma(1) = q \right\}.
\end{equation}
%

%%%%%%%%%%%%%%%%%%%%%%%%%%%%%%%%%%%%%%%%%%%%%%%%%%%%%%%%%%%%%%%%%%%%%%%%%%%%%%%%
\section{Proper latent decomposition} \label{sec:methodology:proper_latent_decomposition}
%%%%%%%%%%%%%%%%%%%%%%%%%%%%%%%%%%%%%%%%%%%%%%%%%%%%%%%%%%%%%%%%%%%%%%%%%%%%%%%%

\begin{figure}
\centering
	\includegraphics[width=\textwidth]{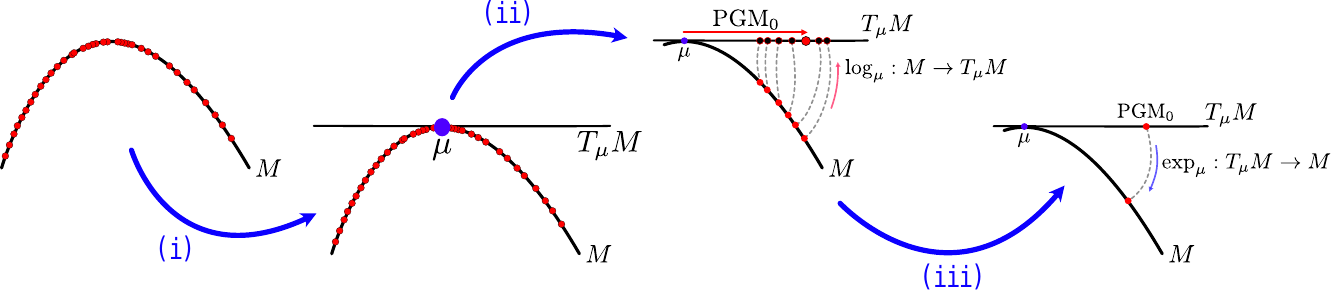}
	\caption{Overview of proper latent decomposition. Given a manifold $M$ (one-dimensional here for pictorial purposes) and data $\mathcal{Z} \subset M$ (red dots), we wish to find modes, or principal geodesics, that best describe this data. This decomposition comprises three stages: (i) a mean $\mu \in M$ is computed on the manifold; (ii) data is mapped to the tangent space centered at the mean, $T_\mu M$, and singular value decomposition is performed to obtain an orthonormal basis in the tangent space; (iii) these basis vectors are mapped back down to the manifold, yielding principal geodesics.}
\label{fig:proper_latent_decomposition}
\end{figure}

Given a Riemannian manifold $(M, g)$, and data $\mathcal{Z} \subset M$ on the manifold, we wish to identify dominant modes on the manifold for both reduced-order modeling and interpreting the underlying manifold. We propose the proper latent decomposition (PLD)~\citep{magri2022InterpretabilityProperLatent} as a means of generalizing linear approaches, such as POD, to nonlinear manifolds. Upon obtaining a suitable nonlinear manifold, the proposed methodology consists of three overarching stages:
(i) computation of the mean on the manifold;
(ii) mapping data from the manifold to the tangent space centered at the mean, and performing singular value decomposition on the tangent space to obtain orthonormal, energy-ordered modes, which describe the underlying data; and 
(iii) mapped the principal components back  to the manifold, which yields the principal geodesics. 
An overview of this process is provided in Figure \ref{fig:proper_latent_decomposition}.

We first identify the nonlinear manifold, which best represents the data through the use of a nonlinear autoencoder. Given high-dimensional data, we can map these to coordinates on the manifold using the encoder; in addition, geometric properties of the manifold, namely the metric tensor, can be inferred by using the decoder. The inferred metric tensor is then used to operate directly on the manifold. Details on the autoencoding approach are provided in Section \ref{sec:methodology:autoencoders}.
Given encoded samples $\mathcal{Z} \subset M$ on the manifold, we  first compute the mean. The arithmetic mean is  incompatible with nonlinear manifolds, being suitable for operating in Euclidean spaces only~\citep{pennec2006IntrinsicStatisticsRiemannian, guigui2023IntroductionRiemannianGeometry}. The Fr\'echet mean accounts for this and is defined as the point on the manifold that minimizes the squared distance to samples on the manifold~\citep{pennec2006IntrinsicStatisticsRiemannian},
\begin{equation} \label{eqn:frechet_mean}
  \mu^\ast = \argmin_{\mu} \sum_{q \in \mathcal{Z}} d_g (\mu, q)^2,
\end{equation}
where $d_g: M \times M \rightarrow \mathbb{R}$ is the distance function on the manifold. The robust computation of the distances between points provides a computational challenge, as explained in Section \ref{sec:review:geodesics}. We make use of the methodology developed in \citet{kelshaw2024ComputingDistancesMeansb} to obtain a continuous, differentiable representation of the distance function, providing an outline of this approach in Section~\ref{sec:methodology:distances}.

Upon identifying the Fr\'echet mean, $\mu \in M$, samples $\mathcal{Z} \subset M$ on the manifold must be mapped to the tangent space defined at the mean using the $\log_\mu: M \rightarrow T_\mu M$ map. Numerical computation of the $\log_\mu$ map is challenging and requires the solution of a two-point boundary value problem, with guarantees that the solution corresponds to a globally length-minimizing geodesic. We introduce a robust numerical framework to achieve this in Section~\ref{sec:methodology:log_map}.

We represent data on the tangent space $\bar{\mathcal{Z}} \in T_\mu M$ as a real-valued matrix, for which we can compute the singular value decomposition
\begin{equation} \label{eqn:singular_value_decomposition}
	\bar{\mathcal{Z}} = U \Sigma V^\top,
\end{equation}
where $U \in \mathbb{R}^{\lvert \mathcal{Z} \rvert \times \lvert \mathcal{Z} \rvert}$ is a unitary matrix, $\Sigma \in \mathbb{R}^{\lvert \mathcal{Z} \rvert \times \dim{M}}$ is a rectangular diagonal matrix with non-negative singular values along the diagonal, and $V \in \mathbb{R}^{\dim{M} \times \dim{M}}$ is the transpose of a unitary matrix. As $U, V$ are unitary, the columns $U_1, \dots, U_{\lvert \mathcal{Z} \rvert}$ of $U$ and the columns $V_1, \dots, V_{\dim{M}}$ of $V$ yield an orthonormal basis. We use the columns of $V$, scaled by their corresponding singular values, as a basis by which to describe the data. The basis vectors can then be mapped back to the manifold through use of the $\exp_\mu$ map. This mapping produces geodesics on the manifold, which maximize the variance of the data and can be visualized by passing the resulting trajectory through the decoder of the trained autoencoder.

%%%%%%%%%%%%%%%%%%%%%%%%%%%%%%%%%%%%%%%%%%%%%%%%%%%%%%%%%%%%%%%%%%%%%%%%%%%%%%%%
\section{Autoencoders for inferring the nonlinear manifold} \label{sec:methodology:autoencoders}
%%%%%%%%%%%%%%%%%%%%%%%%%%%%%%%%%%%%%%%%%%%%%%%%%%%%%%%%%%%%%%%%%%%%%%%%%%%%%%%%

Consider high-dimensional data $\mathcal{X} \subset \mathbb{R}^n$, which can be expressed on a lower dimensional manifold $M$, for which $\dim{M} \ll n$. We wish to obtain a numerical representation of the underlying manifold, leveraging the expressivity of neural networks~\citep{hornik1989MultilayerFeedforwardNetworks}, in particular autoencoders, to achieve this.  

An autoencoder $\eta_\theta: x \mapsto x$ is a neural network suitable for obtaining a numerical representation of the underlying manifold, comprising an encoder $\mathcal{E}_{\theta_e}: \mathbb{R}^n \rightarrow M$ which seeks to obtain coordinates $z \in M$; and a decoder $\mathcal{D}_{\theta_d}: M \rightarrow \mathbb{R}^n$ which is tasked with reconstructing the original input. For convenience, we set $\theta = \{ \theta_e, \theta_d \} \in R^{p}$. Concretely, the encoder and decoder are composed as
\begin{equation} \label{eqn:autoencoder}
	\eta_{\theta} : x \mapsto \left( \mathcal{D}_{\theta_d} \circ \mathcal{E}_{\theta_e} \right)(x).
\end{equation}
The encoder $\mathcal{E}_{\theta_e}$ and decoder $\mathcal{D}_{\theta_d}$  are represented as a composition of multiple linear layers, each followed by an element-wise nonlinear activation function $\sigma$. In the absence of these nonlinear activations, the network is only capable of learning a linear transformation, which restricts the function approximation space.
In this work, we deal exclusively with structured data on grids, allowing us to make modeling choices and construct the encoder and decoder with a series of convolutional layers. These convolutional layers are translation equivariant and allow us to hard-constrain further information, including boundary conditions. To obtain a numerical representation of the manifold, we seek parameters $\theta$, which minimize the mean-squared error of the reconstructions.

%%%%%%%%%%%%%%%%%%%%%%%%%%%%%%%%%%%%%%%%%%%%%%%%%%%%%%%%%%%%%%%%%%%%%%%%%%%%%%%%
\section{Computation of distances} \label{sec:methodology:distances}
%%%%%%%%%%%%%%%%%%%%%%%%%%%%%%%%%%%%%%%%%%%%%%%%%%%%%%%%%%%%%%%%%%%%%%%%%%%%%%%%

In previous work~\citep{kelshaw2024ComputingDistancesMeansb}, we proposed a methodology for obtaining numerical representations of distance functions directly on the manifold. Distance functions are solutions to the Eikonal equation, for which the magnitude of the gradient of the solution is unity at all points in the domain. On a Riemannian manifold $(M, g)$, we require a more general definition of the gradient: The gradient $\nabla \varphi$ of a function $\varphi: M \rightarrow \mathbb{R}$ is the unique vector in a vector space $V$ such that the inner product with any element of $V$ is the directional derivative of $\varphi$ along the vector; that is,
\begin{equation}
	\langle \nabla \varphi, \cdot \rangle_g = d\varphi = \frac{\partial \varphi}{\partial x^i} \partial^i,
\qquad \text{and so,} \qquad
	\nabla \varphi = g^{ij} \frac{\partial \varphi}{\partial x^i} \partial_j,
\end{equation} 
where $g^{ij}$ is the inverse of the metric tensor and $\partial^i, \partial_j$ are the local covariant and contravariant bases, respectively. Using this definition, we can express the Eikonal equation
\begin{equation}
	d\varphi(\nabla \varphi) = \langle \nabla \varphi, \nabla \varphi \rangle_g = 1
\quad \text{s.t.} \quad
	\restr{\varphi}{p} = 0.
\end{equation}
Given two points on the manifold, the length-minimizing path must travel orthogonal to level-sets of distance. Therefore, the gradient of the distance function satisfies the geodesic equation
\begin{equation} \label{eqn:distance:geodesics}
	\nabla_{\nabla \varphi} \nabla \varphi = 0.
\end{equation}
Although we provide a pedagogical overview here for the Eikonal equation as a function of one variable, we defer the extension to multiple variables, as well as obtaining a numerical representation, to~\citet{kelshaw2024ComputingDistancesMeansb}.

%%%%%%%%%%%%%%%%%%%%%%%%%%%%%%%%%%%%%%%%%%%%%%%%%%%%%%%%%%%%%%%%%%%%%%%%%%%%%%%%
\section{Robust computation of $\log_p$ maps} \label{sec:methodology:log_map}
%%%%%%%%%%%%%%%%%%%%%%%%%%%%%%%%%%%%%%%%%%%%%%%%%%%%%%%%%%%%%%%%%%%%%%%%%%%%%%%%

Given points $p, q \in M$, the $\log_p$ map finds a tangent vector $v \in T_p M$, for which $\exp_p v = q$. There might be multiple solutions, yet it is only the length-minimizing solution we seek. Computing the $\log_p$ map means solving a two-point boundary value problem for the geodesic equation; that is,
\begin{equation} \label{eqn:boundary_value_problem}
\text{solve} \quad
	\nabla_{\dot{\gamma}} \dot{\gamma} = 0
\quad \text{for } \dot{\gamma}(\lambda = 0) \quad \text{subject to} \quad 
	\restr{\gamma}{\lambda = 0} = p, \; 
	\restr{\gamma}{\lambda = 1} = q.
\end{equation}
A standard approach for computing the $\log_p$ map is to use a shooting method. Although shooting methods yield accurate results, they rely on a good initial guess. In cases where a sufficiently good approximation of the solution is not available, there are no convergence guarantees. To address this, we develop a framework for numerical computation of the $\log_p$ map that employs the learned distance function on the manifold, leveraging the fact that the gradient of the distance function can be used to obtain length-minimizing curves, as shown in Eq.~\eqref{eqn:distance:geodesics}.
For each point $p \in M$, we can restrict the distance function such that $d_{g|p} : q \mapsto d_g (p, q)$ for all $q \in M$, allowing us to compute trajectories of the form
\begin{equation}
	\tilde{\gamma}_{a \rightarrow b}(\lambda) = a - d_{g|b} (a) \int_0^1 \nabla d_{g|b} \circ \tilde{\gamma}_{a \rightarrow b}(\lambda) d\lambda, 
		\qquad \text{where } \restr{\tilde{\gamma}_{a \rightarrow b}}{\lambda = 0} = a,
\end{equation}
by integrating in the direction of the gradient of the distance field. When the true distance function is known, this will yield the true length-minimizing geodesic. When we only have an approximation of the distance function, as is the case with the numerical representation of the solution to the Eikonal equation, we can use these trajectories to produce reasonable estimates of the geodesics. Given points $p, q \in M$, a pair of trajectories, each with a known boundary condition, can be generated and interpolated between,
\begin{equation} \label{eqn:interpolating_geodesic}
	\gamma_{pq}(\lambda) = 
		(1 - \lambda) \tilde{\gamma}_{p \rightarrow q}(\lambda) 
		+ \lambda \tilde{\gamma}_{q \rightarrow p}(1 - \lambda),
	\qquad \text{s.t. }
		\restr{\gamma_{p \rightarrow q}}{\lambda = 0} = p,
		\restr{\gamma_{p \rightarrow q}}{\lambda = 1} = q,
\end{equation}
producing an approximation of the length-minimizing geodesic. With this approximate geodesic, we use a direct multiple shooting approach to further refine the geodesic. Direct multiple shooting methods provide improved numerical stability when compared to standard shooting methods, and allow for parallel evaluation over sub-intervals, yielding performance improvements in proportion to the number of intervals~\citep{press2007NumericalRecipes3rd}. On reaching convergence, the standard shooting method is used to refine the solution, typically requiring only $\mathcal{O}(1)$ evaluations to reach machine precision. Both the direct multiple shooting and standard shooting methods make use of Gauss-Newton iterations to reach convergence. In this work, the sensitivity of the residual with respect to parameters is computed using a backward-in-time, continuous adjoint approach,  which reduces memory requirements  when compared to employing the standard discretize-then-optimize approach used in conventional automatic differentiation~\citep{chen2019NeuralOrdinaryDifferentiala}.

%%%%%%%%%%%%%%%%%%%%%%%%%%%%%%%%%%%%%%%%%%%%%%%%%%%%%%%%%%%%%%%%%%%%%%%%%%%%%%%%
\section{Results} \label{sec:results}
%%%%%%%%%%%%%%%%%%%%%%%%%%%%%%%%%%%%%%%%%%%%%%%%%%%%%%%%%%%%%%%%%%%%%%%%%%%%%%%%
We show results for two systems of interest: a laminar flow past a bluff body and the two-dimensional turbulent Kolmogorov flow.

\subsection{Laminar flow trailing a bluff body} \label{sec:results:laminar}
We first consider the laminar wake past a triangular bluff body at $Re = 100$. A data set of $800$ snapshots of vorticity in a box downstream of the body is generated at a time interval of $\Delta t = 0.1 s$. Because the dynamics are periodic, we train a convolutional autoencoder with  two latent variables, $\dim M = 2$. We employ the Adam optimizer \citep{kingma2017AdamMethodStochastic} with a learning rate of $\eta = 3 \times 10^{-4}$ to perform  gradient-based optimization for a total of $5 \times 10^{4}$ parameter updates. The trained autoencoder achieves a relative $\ell^2$ error of $1.378 \times 10^{-2}$. 
Next, a numerical representation of the distance function on the manifold is obtained by training a network as described in~\citet{kelshaw2024ComputingDistancesMeansb} and \citet{magri2022InterpretabilityProperLatent} using the metric inferred by the trained autoencoder. The same optimization configuration is employed, with each parameter update consisting of $2^{12}$ pairs of coordinates, drawn random-uniformly across the domain. The obtained distance function achieves a mean residual of $1.734 \times 10^{-3}$ across the domain after $10^5$ parameter updates. This trained distance function is  used to compute the Fr\'echet mean on the manifold, using gradient-descent with a step size of $\eta = 10^{-1}$.

Adopting the framework outlined in Section~\ref{sec:methodology:log_map}, we map the encoded data on the manifold to the tangent space defined at the mean in three stages. First, geodesic interpolants are computed as shown in Eq.~\eqref{eqn:interpolating_geodesic}, using a fourth-order explicit Runge-Kutta scheme with a step size of $\Delta \lambda = 10^{-2}$. The resulting trajectory is broken down into eight sub-intervals, providing an initial guess for the multiple shooting solver. The multiple shooting solver uses the continuous adjoint of the geodesic equation to refine the solution until the $L_2$ norm of the Gauss-Newton update falls below a tolerance of $10^{-4}$, using the same numerical integration scheme. The standard shooting solver is used to optimize  the initial condition in a similar manner. Finally, the singular value decomposition of the data on the tangent space at the mean is carried out, and the scaled basis vectors are mapped back to the manifold using the $\exp_\mu$ map.
\begin{figure}
\centering
	\includegraphics[width=\textwidth]{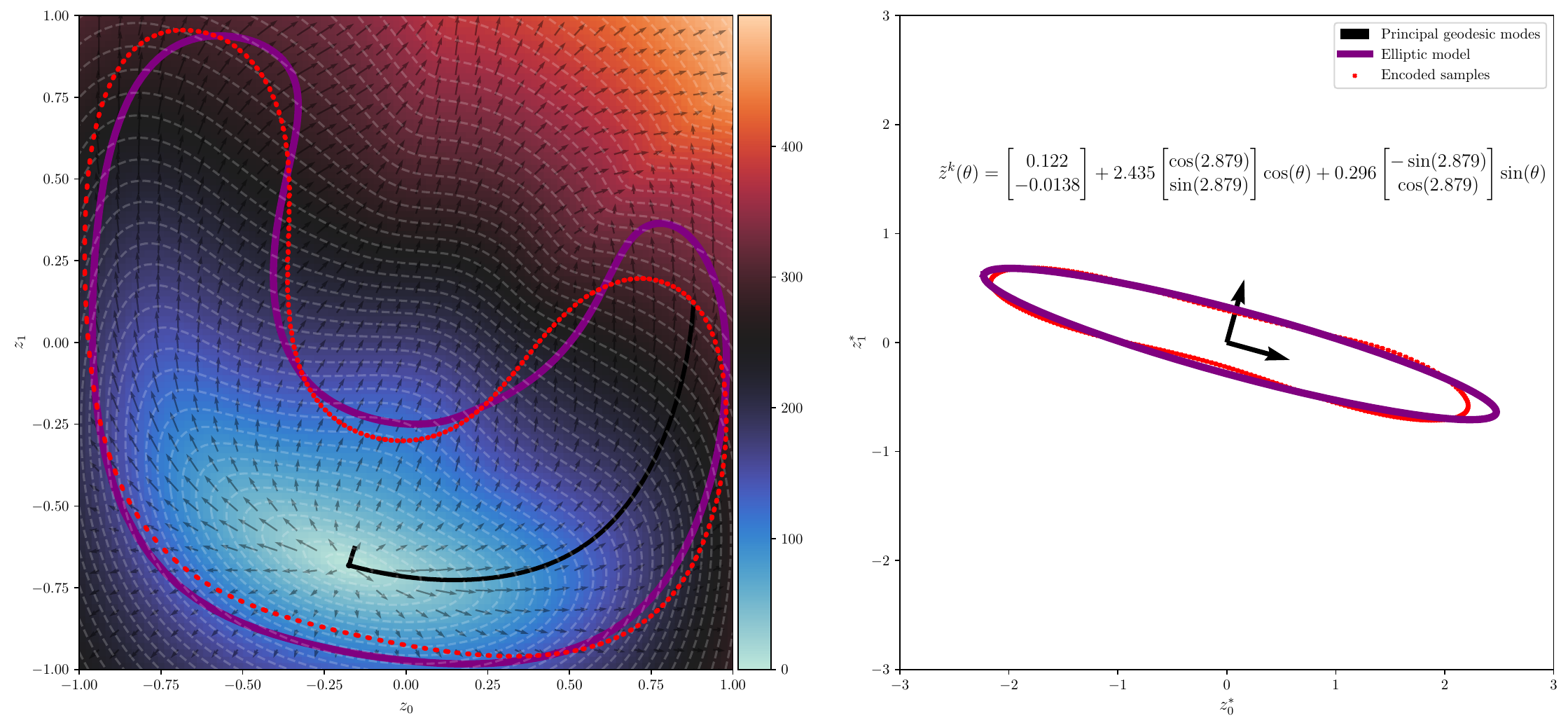}
	\caption{Proper latent decomposition of the laminar flow. Left and right panels show the manifold and the tangent space to the mean respectively. Encoded snapshots of vorticity are depicted as red points, and the resulting geodesic modes are shown in black. An ellipse has been fit to the data in the tangent space, and the resulting trajectory is shown in purple in both panels. The left panel displays contours of geodesic distance around the Fr\'echet mean, highlighting the nonlinearity of the manifold.}
	\label{fig:results:laminar:manifold}
\end{figure}

\begin{figure}
\centering
	\includegraphics[width=\textwidth]{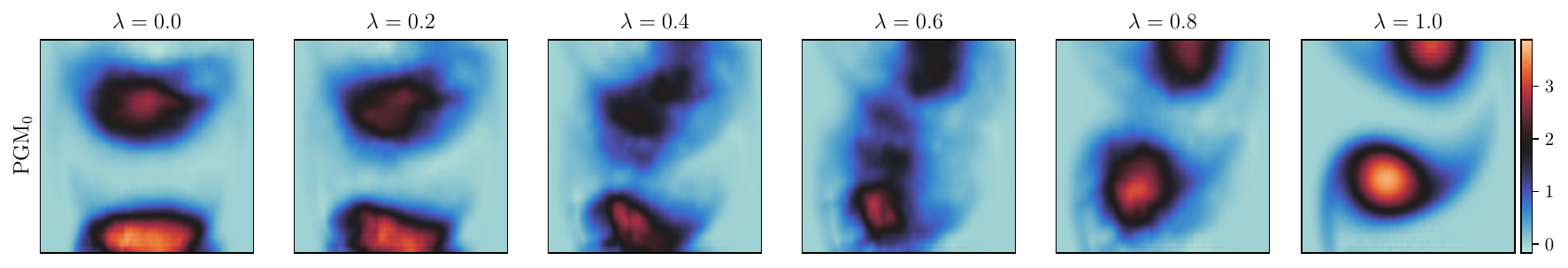}
	\caption{Leading principal geodesic mode for the laminar wake. Snapshots are visualized along the trajectory of the principal geodesic, showing the transition from the Fr\'echet mean, to a physical snapshot of the laminar wake.}
\label{fig:results:laminar:pgm}
\end{figure}

Results are shown in Figure~\ref{fig:results:laminar:manifold}. The left panel of Figure~\ref{fig:results:laminar:manifold} shows the encoded samples, the learned distance field and geodesic flow around the Fr\'echet mean, as well as the principal geodesic modes. The right panel of Figure~\ref{fig:results:laminar:manifold} shows the encoded data mapped onto the manifold as well as the leading principal mode. Because of the structure of the data in the tangent space, we can fit a reduced-order model in the tangent space in the form of an ellipse, describing the entire flow field with a single variable, $\theta$.
The resulting principal geodesic modes are inputted into the trained decoder to be visualized in Figure~\ref{fig:results:laminar:pgm}. We observe the transition from the state at the Fr\'echet mean to that of a snapshot describing the turbulent flow. The structure of the trailing vortices is captured, which shows that the principal modes are physical. 

\subsection{Kolmogorov flow} \label{sec:results:kolmogorov}

We consider the turbulent Kolmogorov flow, which is a two-dimensional incompressible flow with fully periodic boundary conditions and a forcing term  to sustain turbulence. A trajectory is simulated at $Re = 34$ to ensure chaotic behavior, with samples recorded at a time-step of $\Delta t = 0.1$. A total of $2^{15}$ samples are used for training, with a further $2^{13}$ used for validation. The manifold describing turbulent flows is inherently more nonlinear than that describing the laminar flow case, and as a result provides a  challenge for the methodology. For visualization purposes, we limit the dimensionality of the latent space to $\dim M = 2$. In addition, we introduce a regularization term in the autoencoder loss to promote smoothness and improve numerical stability; this is discussed further in Section~\ref{sec:discussion}. The autoencoder achieves a relative $\ell^2$ error of $1.59 \times 10^{-2}$ on the validation set.

\begin{figure}
\centering
	\includegraphics[width=\textwidth]{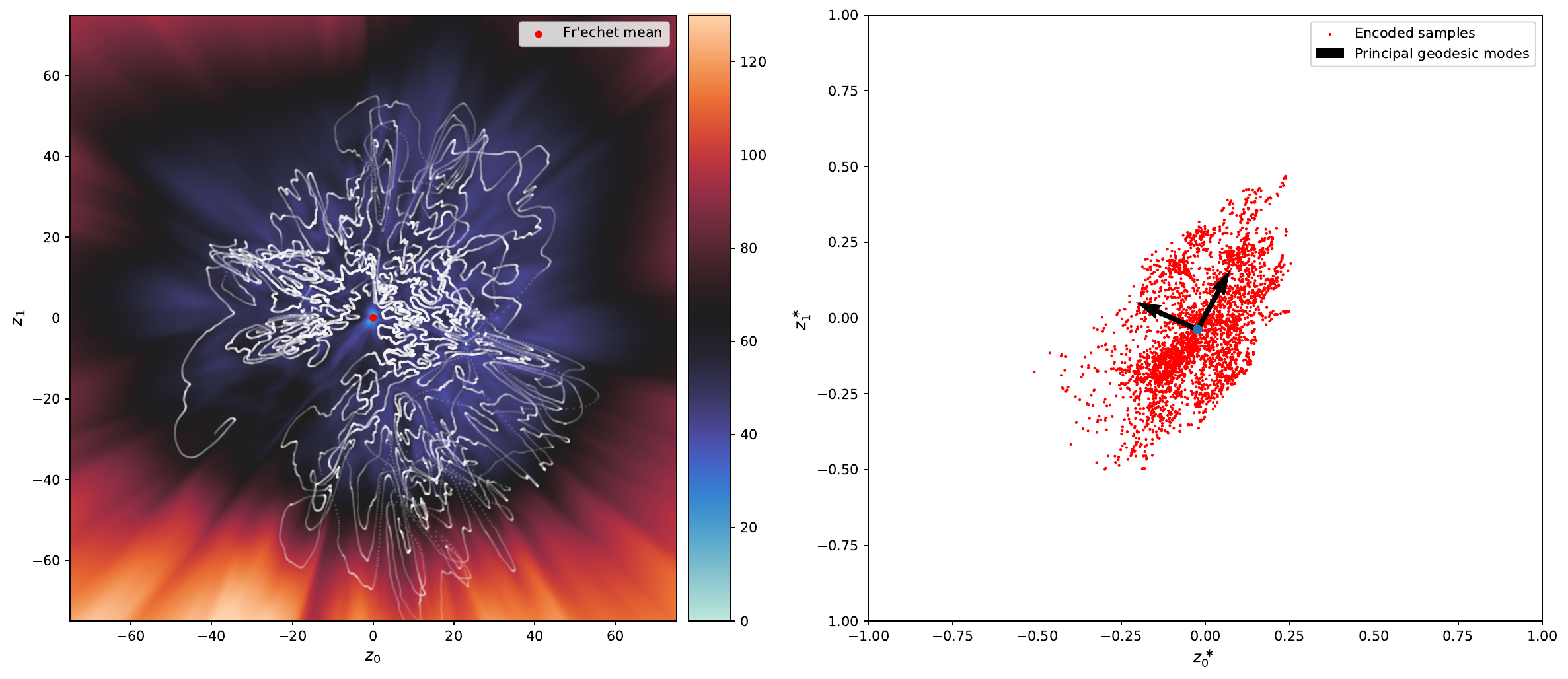}
	\caption{Proper latent decomposition of the Kolmogorov flow. Left and right panels show the manifold and the tangent space to the mean respectively. Encoded snapshots of vorticity are depicted as white points on the manifold, with the colormap showing the distance from the Fr\'echet mean.}
	\label{fig:results:kolmogorov:manifold}
\end{figure}

\begin{figure}
\centering
	\includegraphics[width=\textwidth]{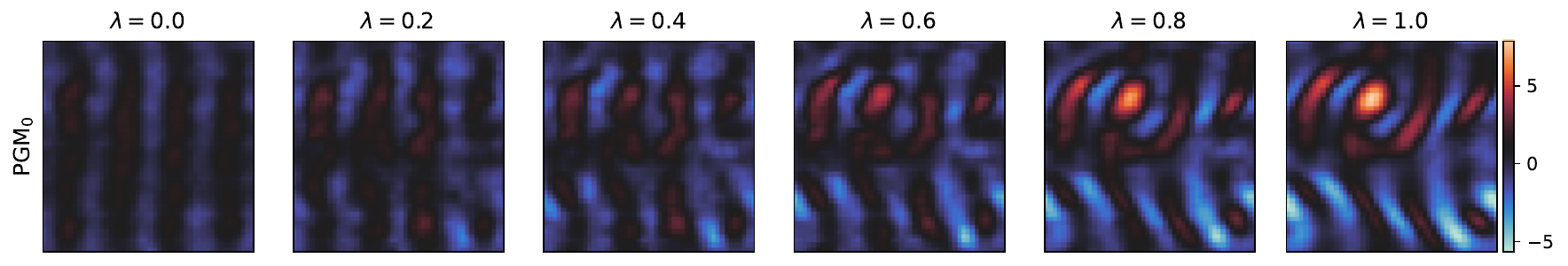}
	\caption{Leading principal geodesic mode for the Kolmogorov flow. Snapshots are visualized along the trajectory of the principal geodesic, showing the transition from the Fr\'echet mean to a physical snapshot of turbulence.}
	\label{fig:results:kolmogorov:pgm}
\end{figure}

For performing PLD, we adopt the same process as described in the laminar flow case (Section~\ref{sec:results:laminar}). The manifold and tangent space at the Fr\'echet mean are shown in Figure~\ref{fig:results:kolmogorov:manifold}, where we observe that encoded samples  on the manifold are  less regular than those seen in the laminar flow case. However, in mapping the samples to the tangent space defined at the mean, we observe a distinct clustering, amenable to singular value decomposition.
Upon obtaining an orthonormal basis, we re-map the scaled basis vectors to the manifold. The trajectory of the leading mode is decoded, with a visualization of the leading mode provided in Figure~\ref{fig:results:kolmogorov:pgm}. We observe a transition from the Fr\'echet mean, which resembles the forcing term in the Kolmogorov flow, to a snapshot of the flow.

%%%%%%%%%%%%%%%%%%%%%%%%%%%%%%%%%%%%%%%%%%%%%%%%%%%%%%%%%%%%%%%%%%%%%%%%%%%%%%%%
\subsection{Discussion} \label{sec:discussion}
%%%%%%%%%%%%%%%%%%%%%%%%%%%%%%%%%%%%%%%%%%%%%%%%%%%%%%%%%%%%%%%%%%%%%%%%%%%%%%%%

\begin{figure}
\centering
	\includegraphics[width=\textwidth]{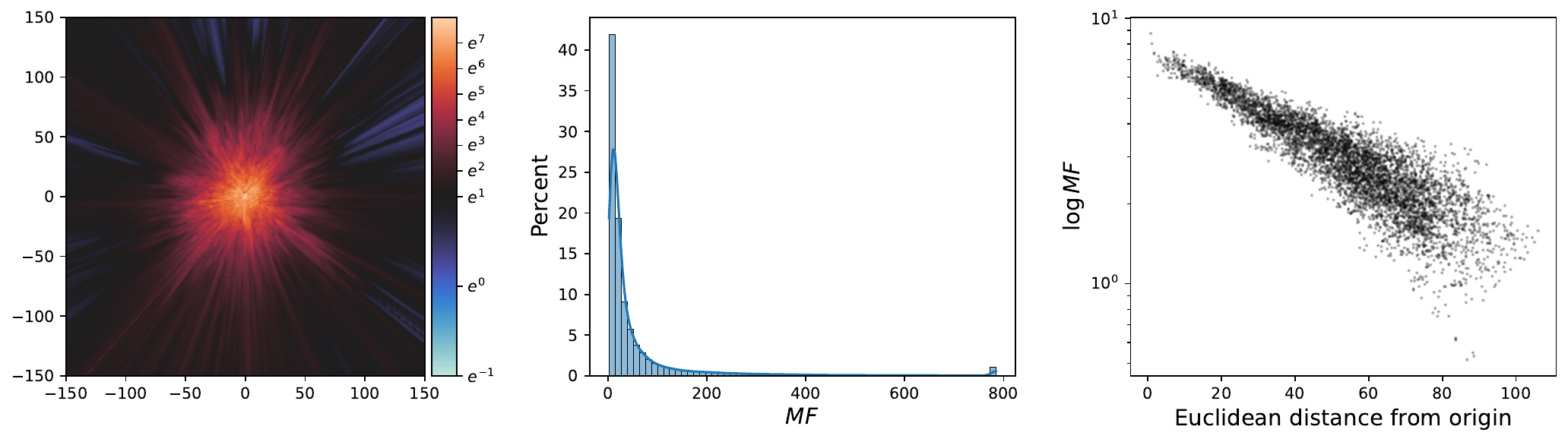}
	\caption{Magnification factor of a prototypical autoencoder trained on the Kolmogorov flow dataset. The left panel demonstrates the $\log MF$ across the latent space; the center panel depicts a histogram of the magnification factor, demonstrating a spike at the tail of the distribution; and the right panel shows the magnification factor as a function of Euclidean distance from the origin.}
	\label{fig:limitations}
\end{figure}

Robust computation of PLD requires the metric tensor inferred from the trained autoencoder to have properties amenable for distance computations. One challenge is that standard optimization methods for autoencoders do not account for these properties, and as a result the numerics can become challenging. To provide a means to inspect the numeric properties of the inferred metric, we analyze the magnification factor, $MF = \sqrt{\det{g}}$, which is a measure of compression at a particular point in space. For prototypical autoencoder trainings, we observe an exponential increase in magnification factor in the center of the domain, often corresponding to metrics whose condition number is exceedingly large. An example of such a manifold is shown in Figure~\ref{fig:limitations}, where the $MF$ is analyzed across the manifold.
Such properties of the metric are a numerical challenge for the analysis and integration of geodesics. To improve numerical stability of the metric across the domain, we introduce regularization terms in the loss function of the autoencoder to promote stability, as proposed by \citet{nazari2023GeometricAutoencodersWhata}. 

%%%%%%%%%%%%%%%%%%%%%%%%%%%%%%%%%%%%%%%%%%%%%%%%%%%%%%%%%%%%%%%%%%%%%%%%%%%%%%%%
\section{Conclusions} \label{sec:conclusions}
%%%%%%%%%%%%%%%%%%%%%%%%%%%%%%%%%%%%%%%%%%%%%%%%%%%%%%%%%%%%%%%%%%%%%%%%%%%%%%%%

In this work, we propose a method for conducting proper latent decomposition~\citep[PLD;][]{magri2022InterpretabilityProperLatent} --- a nonlinear framework for reduced-order modeling. This method identifies a low-dimensional set of intrinsic coordinates using an autoencoding approach before employing tools from differential geometry to analyze and exploit the inferred nonlinear manifold. Generalizing  proper orthogonal decomposition, PLD  allows us to compute principal geodesics on the manifold, providing, for the first time, a manifold-aware approach to nonlinear reduced-order modeling. We demonstrate results for two challenging cases, the laminar wake past a triangular bluff body and the two-dimensional Kolmogorov flow. In the case of the laminar wake, we are able to obtain principal geodesics, which describe the wake dynamics as well as produce a semi-analytical solution of the Navier-Stokes equations by leveraging the geometry of the attractor. In the Kolmogorov flow case, we discuss challenges in identifying intrinsic coordinates amenable to proper latent decomposition. We remedy these issues through geometric regularization of the latent space, allowing us to identify a dominant mode, which, in contrast to POD, is a physical realization of the governing equations. Finally, ongoing challenges and practical considerations are discussed, highlighting areas on which further research will be focused. This work opens opportunities for reduced-order modeling on nonlinear manifolds.

%%%%%%%%%%%%%%%%%%%%%%%%%%%%%%%%%%%%%%%%%%%%%%%%%%%%%%%%%%%%%%%%%%%%%%%%%%%%%%%%
% THE BIBLIOGRAPHY
%%%%%%%%%%%%%%%%%%%%%%%%%%%%%%%%%%%%%%%%%%%%%%%%%%%%%%%%%%%%%%%%%%%%%%%%%%%%%%%%
% \bibliographystyle{ctr}
% \bibliography{references}

\end{document}